\documentclass[preprint,12pt]{elsarticle}
\usepackage{url}

\usepackage{amssymb}
\usepackage{amsmath}
\usepackage{algorithmic}
\usepackage{algorithm}
\usepackage{float}

\journal{Medical Image Analysis}
\usepackage{booktabs}
\usepackage{subcaption}
\usepackage{multirow}
\begin{document}
\begin{frontmatter}



\title{M4: Multi-Proxy Multi-Gate Mixture of Experts Network for Multiple Instance Learning in Histopathology Image Analysis} 


\author[org1]{Junyu Li\fnref{equal}}
\ead{lijunyu1019@gmail.com}
\author[org1,org2]{Ye Zhang\fnref{equal}}
\ead{zhangye22@mails.ucas.ac.cn}
\author[org1,org3]{Wen Shu}
\ead{shuwen@hnu.edu.cn}
\author[org1,org3]{Xiaobing Feng}
\ead{cpufxb@hnu.edu.cn}
\author[org1,org2]{Yingchun Wang}
\ead{wangyingchun22@mails.ucas.ac.cn}
\author[org1]{Pengju Yan}
\ead{yanpengju@gmail.com}
\author[org1]{Xiaolin Li}
\ead{xiaolinli@ieee.org}
\author[org1]{Chulin Sha\corref{corresponding}}
\ead{shachulin@gmail.com}
\author[org1,org2,org3]{Min He\corref{corresponding}}
\ead{hemin607@163.com}

\cortext[corresponding]{Corresponding author.}
\fntext[equal]{The two authors contribute equally to this work.}

\affiliation[org1]{organization={Hangzhou Institute of Medicine (HIM)},
            addressline={Chinese Academy of Sciences}, 
            city={Hangzhou},
            postcode={310022}, 
            country={China}}
\affiliation[org2]{organization={Hangzhou Institute for Advanced Study},
            addressline={University of Chinese Academy of Sciences}, 
            city={Hangzhou},
            postcode={310024}, 
            country={China}}
\affiliation[org3]{organization={College of Electrical and Information Engineering},addressline={Hunan University}, city={Changsha},postcode={410082}, country={China}}

\begin{abstract}
Multiple instance learning (MIL) has been successfully applied for whole slide images (WSIs) analysis in computational pathology, enabling a wide range of prediction tasks  from tumor subtyping to inferring genetic mutations and multi-omics biomarkers. However, existing MIL methods predominantly focus on single-task learning, resulting in not only overall low efficiency but also the overlook of inter-task relatedness. To address these issues, we proposed an adapted architecture of Multi-gate Mixture-of-experts with Multi-proxy for Multiple instance learning (M4), and applied this framework for simultaneous prediction of multiple genetic mutations from WSIs. 
The proposed M4 model has two main innovations: (1) utilizing a mixture of experts with multiple gating strategies for multi-genetic mutation prediction on a single pathological slide; (2) constructing multi-proxy expert network and gate network for comprehensive and effective modeling of pathological image information. Our model achieved significant improvements across five tested TCGA datasets in comparison to current state-of-the-art single-task methods. The code is available at: \url{https://github.com/Bigyehahaha/M4}.
\end{abstract}




\end{frontmatter}



\section{Introduction}
\label{sec:intro}

Advances in digitizing of histopathological whole slide images (WSIs) have significantly heightened the interest in computational pathology (CPath) \cite{shmatko2022artificial,song2023artificial,li2022comprehensive}. Recent deep learning-based WSI analyses have demonstrated substantial progress in automating clinical diagnosis, prognosis and treatment response prediction \cite{zhang2019pathologist,skrede2020deep,li2021deep}. Moreover, due to the important role of molecular ‘omics’  in precision medicine, contemporary CPath has also been used to infer omics features such as microsatellite instability \cite{kather2019deep}, genetic mutations \cite{fu2020pan,qu2021genetic}, and multi-omics aberrations \cite{tsai2023histopathology} among others. These aid in the discovery of new biomarkers and offer a cost-effective alternative compared to complex omics assays. Consequently, CPath holds great potential in decoding different biological phenomena and extracting comprehensive information from a single WSI. This is particularly useful as an increasing number of omics features are proven to be clinically relevant, it becomes impractical to train a deep-learning model for every single task. However, currently, the majority of the WSI analyses typically focus on a specific task, partly due to the difficulties in the complex histopathological phenotype interpretation and the scarcity of available annotations for various tasks. 
\begin{figure}[t]
  \centering
  \includegraphics[width=1\linewidth]{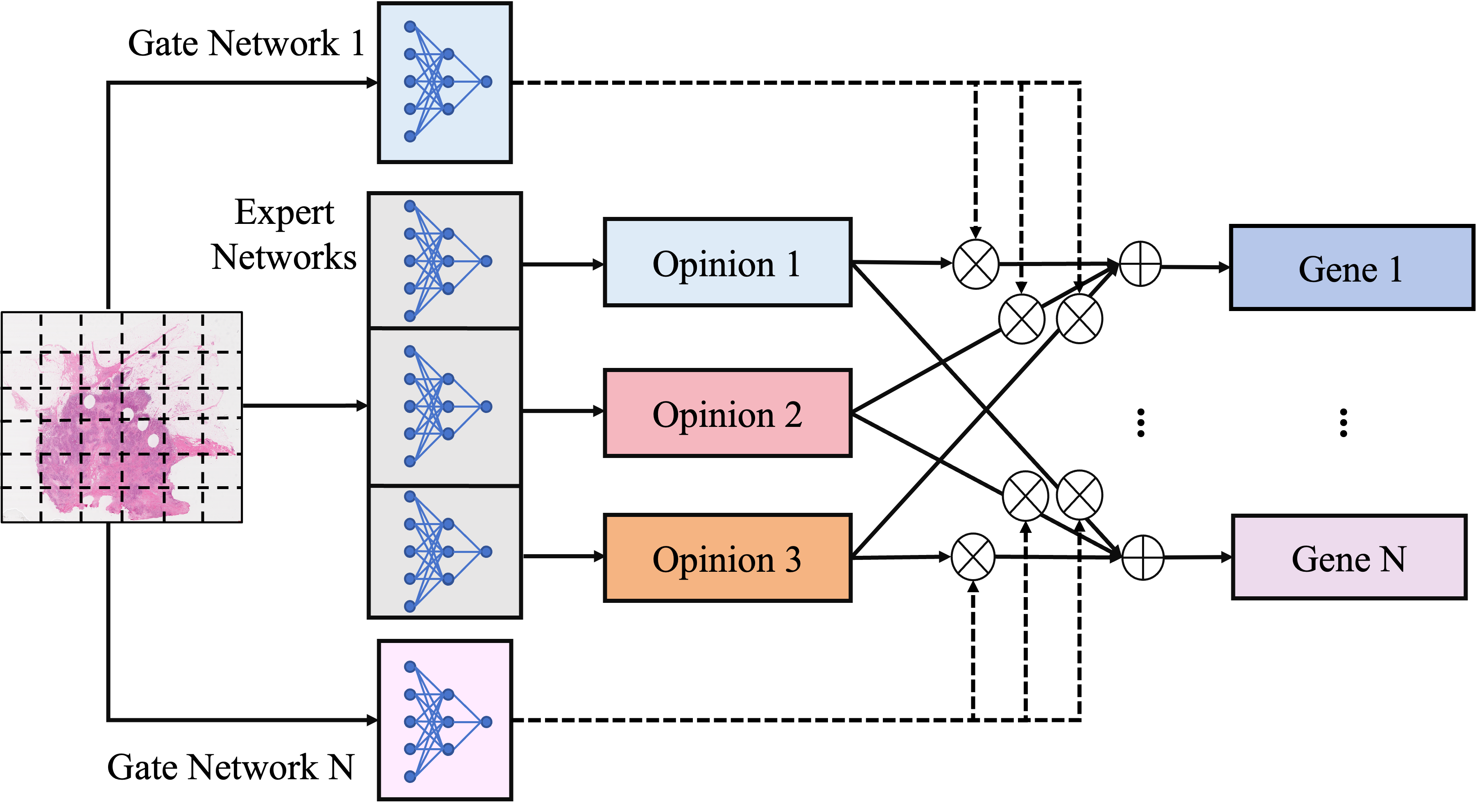}
   \caption{Illustration of multi-task prediction of genetic mutations. Given a WSI, the integrating MMoE and MIL method can be used to simultaneously predict multiple genetic mutations.}
   \label{fig_1}
\end{figure}
There are several main challenges in deep leaning-based WSI analysis. First, compared to natural or small medical images, WSIs are extremely large, ranging from 100M pixels to 10G pixels, making single-pass analysis over the entire slide computationally intractable \cite{abdel2023comprehensive,kumar2020whole}. A common practice to address this problem is patching, wherein WSIs are divided into smaller patches and processed individually. The outputs from these patches are subsequently aggregated to derive a result at the slide level. Another major challenge in WSI analysis is the shortage of annotations, because labelled data is typically only available at the slide-level, and manual annotations on patches or regions of interest by pathologists are both time-consuming and labor-intensive \cite{hu2023state,hou2022h}. Moreover, the annotations become ambiguous for tasks like genetic mutation prediction, further exacerbating the paucity of supervised training data. Recent progress has adopted `weakly-supervised' technique to develop deep learning-based WSI models trained with limited annotations. This is also referred as Multiple Instance Learning (MIL), which assumes only a subset of patches correspond to a slide-level supervisory label \cite{li2022comprehensive,kumar2020whole,kanavati2020weakly}. MIL strategy has gained popularity in WSI analysis and typically includes three stages: 1) Patch-level features extraction from uniform square tiles using a self-supervised pre-trained feature extractor \cite{wang2022scl,wang2021transpath,lazard2023giga}; 2) Global feature aggregation to derive a WSI-level representation \cite{schmidt2022efficient,xie2020beyond}; 3) End-point task prediction based on WSI-level representation. \par

More recent advances in deep learning-based WSI models mainly concentrate on  improving the performance of MIL-based methods such as attention-based MIL \cite{ilse2018attention,chen2022pan} and transformer-integrated MIL \cite{shao2021transmil,li2021dt}, as well as patch-level representation learning through self-supervised learning pre-training networks \cite{lazard2023giga,wang2022scl}. Nonetheless, there are relatively few studies working on multi-task learning for WSIs \cite{gao2023semi,liu2023multi}. However, with the rapid progress in the precision medicine field, an increasing number of omics features are proven of clinical importance and potentially inferable with WSIs, it becomes inefficient and impractical to train a deep learning model for every specific task. Moreover, in the case of omics features such as genetic mutations, which exhibit intricate interactions and co-expression patterns within the tumor microenvironment, the WSI reflects the collective effects of numerous mutations on the histopathology phenotype rather than independent events. Therefore, current single-task approaches for genetic mutation prediction from WSIs suffer from several limitations: 1) The complex interplay and synergistic mutation of genes in WSI are overlooked; 2) Imbalanced positive and negative samples result in overfitting and poor learning capability for infrequently mutated genes; 3) Inefficient inference require separate models with redundant computations for each gene. \par

To address these issues, we investigate the application of multi-task learning for WSI analysis inspired by the Multi-gate Mixture-of-Experts (MMoE) strategy \cite{ma2018modeling,zheng2022survey} (Figure \ref{fig_1}), which  is commonly used in recommendation systems. Specifically, we adapted the MMoE framework with a multi-proxy MIL expert network and gate network for comprehensive and effective WSI information extraction. The experts are aggregated through task-specific gated networks and fed into the corresponding task classifier to achieve mutation prediction. With this approach, the M4 model can: 1) learn shared features beneficial for all gene mutation tasks while permitting unique learning for individual gene mutations; and 2) comprehensively model the relationships between patches at different scales through multi-proxy MIL experts. To the best of our knowledge, we are the first to apply the MMoE architecture to multi-genetic mutation prediction tasks. Compared to existing single-task models, our proposed model improves almost overall performance across multiple gene mutations in various cancer types.\par
We evaluate our M4 framework on five cancer datasets (GBMLGG, BRCA, CRC, UCEC, and LUAD) from The Cancer Genome Atlas (TCGA) to predict the top 10 most frequently mutated genes for each cancer type. Our proposed M4 model consistently outperforms state-of-the-art single-task methods for an average AUC increase of 4\%. Furthermore, we validate the efficacy of our multi-proxy experts MP-AMIL and multi-proxy gate networks MP-Gate across the aforementioned cancer types, both prove to be effective in improving the performance of the model. We also assess how the number of experts and tasks affects the M4 model's performance. In addition, heatmap visualizations of our M4 model and the single-task model on the same pathological slides demonstrate that the M4 model has an enhanced focus on tumor regions, especially for rarely mutated genes.\par
Overall, we design an innovative M4 framework to address the multi-task learning problem in WSI analysis, demonstrating its superiority over state-of-the-art single-task models. This work offers a new type of framework for WSI analysis and is particularly beneficial amid the transition toward precision medicine. \par
\section{Related work}
\subsection{Multiple Instance Learning in WSI Analysis}

In WSI analysis and pixel-level annotations of histopathology images, images of gigapixel-level pose significant challenges.
Recently, several works have demonstrated great potential and achieved remarkable progress with MIL \cite{campanella2019clinical,lerousseau2020weakly,xu2019camel}. MIL learns to map the patches to the labels through patch-embedding aggregation process. Ilse et al. \cite{ilse2018attention} first adopted the attention mechanism with trainable parameters to aggregate patch-level features into WSI-level representations. Chen et al. \cite{chen2022pan} improved on the basis of previous research and proposed attention-based MIL (AMIL). \par
More recently, new methods considering the inter-instance relationships have been developed. Li et al. \cite{li2021dual} introduced a dual-stream architecture-based MIL (DSMIL) to improve the attention mechanism. Shao et al. \cite{shao2021transmil} designed transformer-based MIL (TransMIL), adopting self-attention networks and pyramid positional encodings to learn the interactions between instances. Zhang et al. \cite{zhang2022dtfd} proposed the double-tier feature distillation MIL framework to relieve overfitting. However, these works primarily focus on single-task learning in WSI analysis, failing to leverage the inter-task correlations for efficient WSI modeling especially for the task of predicting gene mutations, which have a high correlation between genes. Moreover, methods based on attention struggle to capture the relationships between patches within a WSI, while Transformer-based methods entail high computational complexity.


\subsection{Multi-task Learning}

Multi-task Learning (MTL) has been extensively applied  in the field of recommendation systems (RS) with significant progress recently \cite{zheng2022survey,tang2020progressive,zhang2022multi}. The major advantage of MTL is that it can exploit rich commonalities across different tasks to improve performance on correlated tasks and enhance model generalization. A commonly used MTL framework is the share-bottom architecture proposed by Caruana\cite{caruana1997multitask}, where the bottom hidden layers are shared across tasks, thus also termed as hard parameter sharing. Reisenbüchler et al. \cite{reisenbuchler2022local} proposed a local attention graph-based Transformer model for WSI analysis with hard parameter sharing for downstream multi-gene mutation prediction. However, these methods may cause negative transfer due to straightforward parameter sharing between conflicting tasks. To address this issue, Jacobs et al. \cite{jacobs1991adaptive} proposed Mixtures of Experts (MoE) where multiple parallel hidden layers termed experts are placed at the model bottom. A gating network is used to combine experts for different tasks. On top of that, Ma J et al. \cite{ma2018modeling} extended MoE by proposing MMoE, where separate gating networks are used to obtain different fusion weights in MTL, thereby capturing the differential information between tasks. Similarly, Zhao et al. \cite{zhao2019multiple} employed multi-head self-attention to learn different representational subspaces over different feature sets for individual tasks. Although MTL has been rapidly evolving in RS, there is still very limited research on MTL for histopathology. \par
In this paper, we adapt the MMoE framework in WSI analysis for multiple gene mutations prediction simultaneously. By testing several approaches to adopt MMoE in MIL architecture, we design a Multi-proxy Multi-gate Mixture-of-experts framework for Multiple instance learning (M4) framework. 


\section{Methods}
In this section, we revisit MIL and MTL, specifically emphasizing the attention-based MIL and MMoE framework.
\subsection{Revisit MIL and MTL  }
\subsubsection{Attention-Based Multiple Instance Learning}
Consider a bag of instances $\mathbf{X}$=$\{\mathit{x_1, x_2, ..., x_N}\}$, where $\mathit{N}$ is the number of instances in the bag. Each instance $\mathit{x_k}$, $\mathit{k}$$\in$ 1, 2, ..., $\mathit{N}$ has a latent label $\mathit{y_k}$ ($\mathit{y_k}=0$ denotes negative, or $\mathit{y_k}=1$ denotes positive). Typically, only the label of the bag is known, while the instance labels are unknown. The primary goal of MIL is to detect whether there exists at least one positive instance in the bag, which is defined as:\par
\begin{equation}
  Y=\left\{
  \begin{array}{ll}
  1, &if \sum_{k=1}^{N}y_{k}>0,\\
  0, &otherwise
  \end{array}
  \right.
  \label{eq:important}
\end{equation}

There are typically two approaches to aggregating instances into a bag. One is the instance pooling strategy, the core idea is to assign each instance the bag label and train a classifier for instance, then apply various pooling operations to obtain the bag prediction \cite{campanella2019clinical,hou2016patch}. The other strategy is bag embedding-based MIL, which means learning the representation of the entire bag from the features of the instances. Ilse et al.\cite{ilse2018attention} demonstrate that bag embedding-based MIL is more effective than the instance pooling strategy. Therefore, in this work, we adopt the bag embedding-based MIL approach, which can be formulated as:
\begin{equation}
  \mathbf{F}=\mathbf{G}(\{x_k|k=1, 2, 3..., \mathbf{N}\})
  \label{eq:important}
\end{equation}
where $\mathit{x_k}\in$$\mathbb{R}^D$ denotes the feature of instance $\mathit{k}$ and $\mathit{D}$ is the dimension of the feature, $\mathbf{G}$ represents the aggregation function that pools the instance-level features into a bag-level feature $\mathbf{F}$. A commonly used and effective aggregation function used in MIL is a simple attention network, defined as:
\begin{equation}
  \mathbf{F}=\sum_{k=1}^{N}\alpha_{k}x_{k} \in\mathbb{R}^D
  \label{eq:important}
\end{equation}
where $\mathit{\alpha_{k}}$ denotes the learnable attention weight for the $\mathit{k}$-th instance,  bag-level feature $\mathbf{F}$ has the same dimension as $\mathit{x_k}$.
\subsubsection{Multi-gate Mixture of Experts Architecture}
Suppose there are $\mathit{n}$ tasks. The core idea of MMoE \cite{ma2018modeling} architecture is to use a mixture of experts' networks while appending a separate gating network $\mathit{g^t}=\{\mathit{g^t_1, g^t_2, ..., g^t_E}\}$ for each task $\mathit{t}$ ($\mathit{t}$$\in$ 1, 2, ..., $\mathit{n}$), where $\mathit{E}$  represents the number of experts. Specifically, the model output of task $\mathit{t}$ is:
\begin{equation}
  y_t=h_t(f_t(x)),
  \label{eq:important}
\end{equation}
\begin{equation}
  f_t(x)=\sum_{e=1}^{E}g^{t}_{e}(x)f_{e}(x),
    \label{eq:important}
\end{equation}
\begin{equation}
  \sum_{e=1}^{E}g^{t}_{e}(x)=1
    \label{eq:important}
\end{equation}
where $\mathit{g^{t}_{e}}$ represents the gating probability of the $\mathit{e}$-th expert $\mathit{f_{e}}$ ($\mathit{e}$=1, 2, 3..., $\mathit{E}$). $\mathit{f_t}$ represents the feature vector obtained by weighted aggregation of all experts' outputs using the gating probabilities from the $\mathit{t}$-th task's gating network $\mathit{g^t}$. Passing $\mathit{f_t}$ through the tower network $\mathit{h_t}$ yields the output $\mathit{y_t}$ for task $\mathit{t}$.\par
\begin{figure*}[!h]
  \centering
  \includegraphics[width=0.95\linewidth]{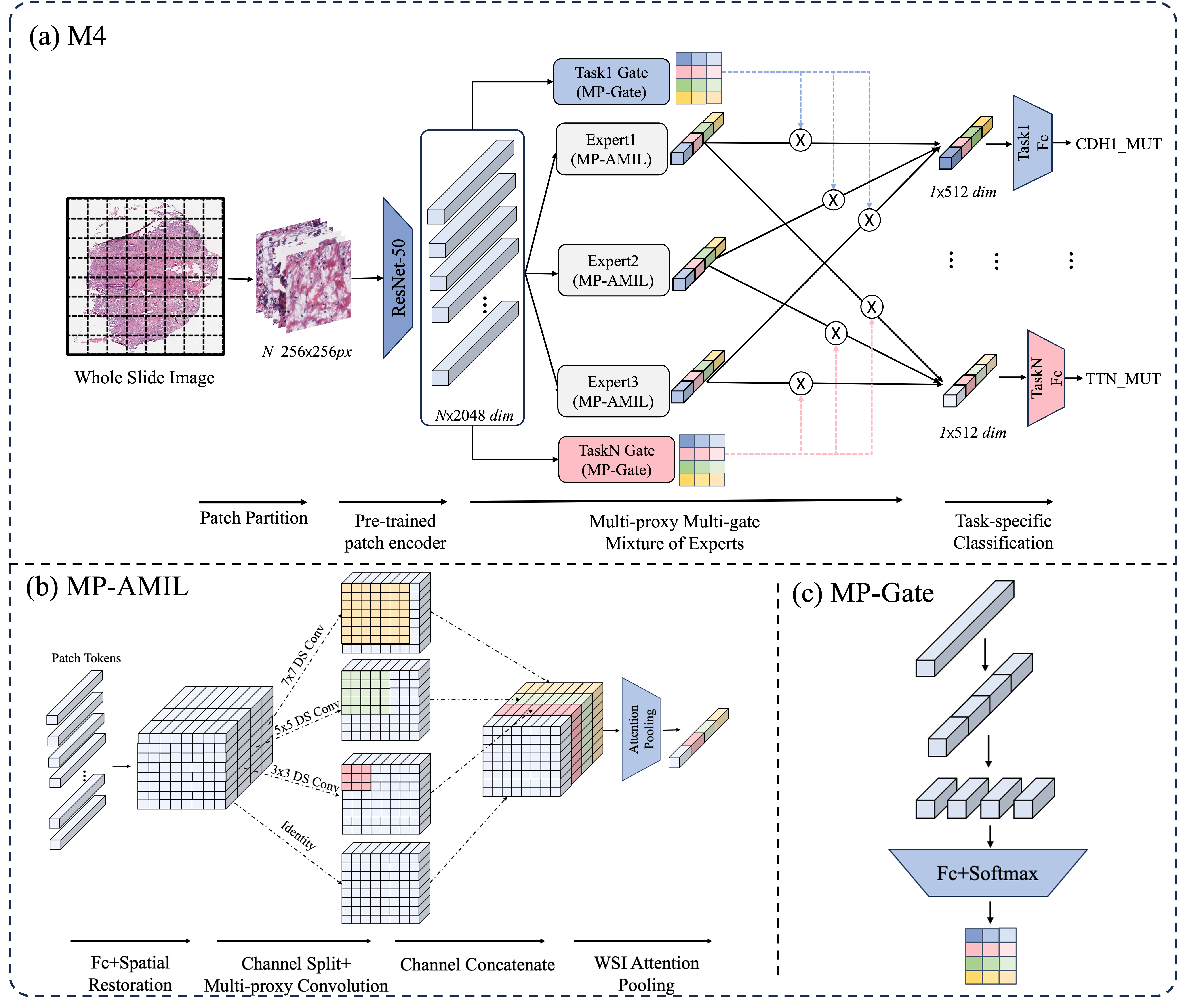}
  \caption{(a)Overview of our proposed M4 architecture. A set of patches are cropped from the tissue regions of a WSI and input to a pre-trained patch-level feature extractor to obtain feature representations of patches. Then all the patches features input to experts and gates layers, and aggregate multi experts information for each task through a multi-proxy MMoE network. Finally, aggregated WSI-level features will be sent to the corresponding tasks' towers for downstream classification. (b)Multi-proxy expert network MP-AMIL. (c)Multi-proxy gate network MP-Gate.}
  \label{fig_2}
\end{figure*}
\subsection{Self-supervised Pre-trained Feature Extraction Models}
In this study, we preprocess each WSI by removing the background, retaining only the tissue regions. The tissue area is then segmented into non-overlapping patches of size $256\times256$ pixels, denoted as $\mathbf{X}$ = $\{\mathit{x_1, x_2, …, x_N}\}$, where $\mathit{N}$ is the total number of patches. We utilize a pre-trained ResNet-50 model specifically trained on pathological images to extract features from these patches \cite{wang2023retccl}. This approach is preferred over using a ResNet-50 model pre-trained on ImageNet, as it is more adept for pathological image analysis. Consequently, each patch is represented by a d-dimensional token, resulting in a representation for the entire WSI as $\mathbf{H}$ = $\{\mathit{h_1, h_2, …, h_N}\}$, where $\mathit{h_i} \in \mathbb{R}^{1\times{d}}$.

\subsection{M4 Model Framework}
In this section, we introduce the proposed M4 framework. The core innovation lies in the design of a specialized MMoE architecture tailored for the MIL framework, as depicted in Figure \ref{fig_2}. We specifically propose a multi-proxy MMoE framework. For the expert component, we construct a computationally efficient multi-proxy multi-instance model to capture the relationships between patches at various scales within pathology images, utilizing an attention network to derive a WSI-level feature representation. For the gating network, we design task-specific gating networks that allocate weights to the multi-proxy experts, resulting in a comprehensive integration that contributes to the final prediction. The subsequent sections will elaborate on each component of our proposed model.
\begin{algorithm}[!h]
    \caption{MP-AMIL Processing Flow}
    \label{alg:AOA}
    \renewcommand{\algorithmicrequire}{\textbf{Input:}}
    \renewcommand{\algorithmicensure}{\textbf{Output:}}
    \begin{algorithmic}[1]
        \REQUIRE A bag of feature tokens $\mathbf{H} \in \mathbb{R}^{N \times d}$;
        \ENSURE WSI-level multi-proxy feature vector $\mathbf{H}_{wsi}$;
        
        \STATE \textbf{Feature Embedding:} Fully connected network with ReLU activation for features embedding; \\
        $\mathbf{H}_{fc} \leftarrow \text{ReLU}(\text{Fc}(\mathbf{H}))$, where $\mathbf{H}_{fc} \in \mathbb{R}^{N \times d_{f}}$;
        
        \STATE \textbf{Spatial Restore:} Reshape patch tokens $\mathbf{H}_{fc}$ to 2D image space; \\
        $\mathbf{H}^{2D}_{fc} \leftarrow \text{Restore}(\mathbf{H}_{fc})$, where $\mathbf{H}^{2D}_{fc} \in \mathbb{R}^{\sqrt{N} \times \sqrt{N} \times d_{f}}$;
        
        \STATE \textbf{Channel Split:} Uniformly divide $\mathbf{H}^{2D}_{fc}$ into 4 segments along the channel dimension; \\
        $\mathbf{H}^{2D,s}_{fc} \leftarrow \mathbf{H}^{2D}_{fc}.\text{split}([d_{f}/4], \text{dim}=-1)$, for $s \in \{1,2,3,4\}$;
        
        \STATE \textbf{Multi-Proxy Convolution:} Apply depthwise separable convolutions (DSC) with different kernel sizes to each segment; \\
        \begin{itemize}
            \item $\mathbf{H}^{2D,1}_{fc} \leftarrow \text{Identity}(\mathbf{H}^{2D,1}_{fc})$;//similar to residual connection
            \item $\mathbf{H}^{2D,2}_{fc,DSC} \leftarrow \text{DSC}(\mathbf{H}^{2D,2}_{fc}, \text{kernel\_size}=3, \text{padding}=1, \text{stride}=1)$;
            \item $\mathbf{H}^{2D,3}_{fc,DSC} \leftarrow \text{DSC}(\mathbf{H}^{2D,3}_{fc}, \text{kernel\_size}=5, \text{padding}=1, \text{stride}=2)$;
            \item $\mathbf{H}^{2D,4}_{fc,DSC} \leftarrow \text{DSC}(\mathbf{H}^{2D,4}_{fc}, \text{kernel\_size}=7, \text{padding}=1, \text{stride}=3)$;
        \end{itemize}
        
        \STATE \textbf{Channel Concatenation:} Concatenate the four segments along the channel dimension; \\
        $\mathbf{H'}^{2D}_{fc} \leftarrow \text{Concatenate}([\mathbf{H}^{2D,1}_{fc}, \mathbf{H}^{2D,2}_{fc,DSC}, \mathbf{H}^{2D,3}_{fc,DSC}, \mathbf{H}^{2D,4}_{fc,DSC}], \text{dim}=-1)$;
        
        \STATE \textbf{Flatten:} Flatten the 2D feature map back to a sequence; \\
        $\mathbf{H'}_{fc} \leftarrow \text{Flatten}(\mathbf{H'}^{2D}_{fc})$, where $\mathbf{H'}_{fc} \in \mathbb{R}^{N \times d_f}$;
        
        \STATE \textbf{Fusion:} Aggregate sequence features using attention pooling to obtain WSI-level feature representations; \\
        $\mathbf{H}_{wsi} \leftarrow \text{Attn-pool}(\mathbf{H'}_{fc})$, where $\mathbf{H}_{wsi} \in \mathbb{R}^{1 \times d_f}$.
    \end{algorithmic}
\end{algorithm}
\subsubsection{Implementation of Experts}
 For given a bag (WSI) of N instances (patches), i.e., $\mathbf{X}$=$\{\mathit{x_1, x_2, ..., x_N}\}$. The feature of patches extracted by a pre-trained encoder can be denoted as $\mathbf{H}$=$\{\mathit{h_1, h_2, ..., h_N}\}$.\par
AMIL\cite{chen2022pan} is a widely used attention-weighted MIL model. It can localize morphological regions that have high task relevance in the WSI, and then aggregate these regions into a single low-dimensional feature representation. Compared to TransMIL, AMIL has lower computational complexity. However, it fails to capture the relationships between different patches in WSI. Therefore, we propose a multi-proxy MIL model(Figure \ref{fig_2}(b)) that employs depthwise separable convolutions\cite{Chollet2017xception} with different kernel sizes across various channels, allowing for a more comprehensive and sufficient modelling of the relationships between patches of different scopes within WSI. This module can enrich the features carried by each patch. Finally, we use an attention mechanism to aggregate patch features, resulting in a WSI-level feature representation.\par
Before the expert module process begins, we perform a squaring operation on the input patch sequences, following the TransMIL model\cite{shao2021transmil}. The processing steps of the expert module are shown in Algorithm \ref{alg:AOA}, where Fc denotes Fully-connected Network, ReLU denotes ReLU activation function, and DSC denotes Depthwise Separable Convolution. Following the acquisition of multi-proxy patch features in step 6, we follow the attention pooling approach in AMIL to aggregate patches features $\mathbf{H'}_{fc}$ into WSI-level feature representations $\mathbf{H}_{wsi}$, which can be formulated as:
\begin{equation}
    \mathbf{H}_{wsi} = \text{Attn-pool}(\mathbf{A}, \mathbf{H'}_{fc}) = \sum_{k=1}^{N} a_k \mathbf{h'}_{k}
\end{equation}
where $\mathbf{A} \in \mathbb{R}^{1 \times N}$ represent the attention scores of all patches in WSI $\mathbf{H'}_{fc}$. Given a patch embedding $h_k \in \mathbb{R}^{1 \times d_{f}}$(the $k$-th patch of $\mathbf{H'}_{fc}$), its attention score $a_k$ is computed by:
\begin{equation}
    a_k = \frac{\exp \left\{ \mathbf{W}_a \left( \tanh \left( \mathbf{V}_a \mathbf{h}_k^\top \right) \odot \text{sigmoid} \left( \mathbf{U}_a \mathbf{h}_k^\top \right) \right) \right\}}{\sum_{k=1}^{N} \exp \left\{ \mathbf{W}_a \left( \tanh \left( \mathbf{V}_a \mathbf{h}_k^\top \right) \odot \text{sigmoid} \left( \mathbf{U}_a \mathbf{h}_k^\top \right) \right) \right\}}
\end{equation}
where $\mathbf{W}_a,\mathbf{V}_a,\mathbf{U}_a$ are the learnable parameters.

\subsubsection{Implementation of The Gates}
In the original MMoE framework, each task is assigned a specific gating network, and each gate network linearly partitions the input space into $\mathit{E}$ (the number of experts) regions, where each region corresponds to a different expert, allowing the learning of weights for assigning different experts. Since our constructed expert network incorporates multi-proxy information, the gate network needs to be correspondingly adapted. In our model, the gate network comprises two fully connected layers followed by a softmax layer. \par
Taking the gate $\mathbf{G}_t$ corresponding to task $t$ as an example, for the WSI feature $\mathbf{H}=(h_1,h_2,...,h_N)$ extracted through a pre-trained encoder, we first employ mean pooling to aggregate all the patch features in $\mathbf{H}$. \\
\begin{equation}
  \mathbf{\Bar{H}} = \frac{1}{N}\sum_{i=1}^{N}h_i
\end{equation}
where $N$ denotes the number of patches in WSI. This aggregated feature is then passed through a fully connected layer for feature mapping and dimension reduction. \\
\begin{equation}
  \mathbf{\hat{H}} = ReLU(\mathbf{W}_{t,1}\mathbf{\Bar{H}})
\end{equation}
where $\mathbf{W}_{t,1}\in$$\mathbb{R}^{d\times{d_1}}$ denotes the weight of the first fully connected layer in gate $t$. The reduced feature is uniformly split into four segments based on channel dimension, corresponding to the multi-proxy expert network. \\
\begin{equation}
  \mathbf{\hat{H}}_1,\mathbf{\hat{H}}_2,\mathbf{\hat{H}}_3,\mathbf{\hat{H}}_4 = \mathbf{\hat{H}}.Split([split\_dim]\times 4,dim=-1)
\end{equation}

Each segment is subsequently processed by another fully connected layer and a softmax layer to obtain the weighted coefficients for all experts specific to task $n$ .
\begin{align}
 gate_i &= Softmax(\mathbf{W}_{t,2}\mathbf{\hat{H}}_i), i =1,2,3,4,\\
 \mathbf{G}_t &=\{gate_1,gate_2,gate_3,gate_4\}.
  \label{eq:gate}
\end{align}
where $\mathbf{W}_{t,2}\in \mathbb{R}^{(d_1/4) \times E}$ denotes the weight of the second fully connected layer in gate $t$ and $\mathbf{G}_t$ denotes the output of gate $t$.

\subsubsection{Implementation of Task Specific Towers}
Just like in the original MMoE framework, the features input to the task-specific tower for task  $t$  can be formulated as:
\begin{equation}
  \mathbf{H}^{in}_t = \sum_{e=1}^E\mathbf{G}_{t,e}\cdot \mathbf{H}_{wsi,e}
  \label{eq:tower_input}
\end{equation}
where $E$ denotes the number of experts. In our model, since both experts and gated networks have multiple segments (different segments contain WSI information of different proxies), the above operation will perform on each segment to obtain WSI weighted aggregation features containing multi-proxy information. After weighing the WSI-level features, we obtain features for the final classification prediction. Subsequently, we apply task-specific MLP heads as towers to predict task labels for each task:
\begin{equation}
  \hat{y}_{1:n}=MLP_{1:n}(\mathbf{H}_{t=1:n}^{in})
  \label{eq:important}
\end{equation}
where $n$ represents the number of tasks.\par
To train the network, we initially employed binary cross-entropy loss for each prediction task. The sum of the losses from all prediction tasks was taken as the final loss:
\begin{equation}
  \mathcal{L}(\sigma)=\sum_{i=1}^{n}-\frac{1}{n}(\sigma(y_{i}log\hat{y}_{i}+(1-y_{i})log(1-\hat{y}_{i})))
  \label{eq:important}
\end{equation}
where $\mathit{y_i}$ denotes the true task label.

\section{Experiments}

\subsection{Dataset Description and Evaluation Details}
We obtained the datasets of cancer diagnostic WSIs and mutation information from The Cancer Genome Atlas (TCGA), which includes: \par
1. Hematoxylin and eosin (H\&E) WSIs.\par
2. The mutation status of top-10 genes with the highest mutation counts.\par
The WSI data was downloaded from the Genomic Data Commons data portal (\url{https://portal.gdc.cancer.gov/}), and mutation labels were sourced from cBioPortal (\url{https://www.cbioportal.org/}).\par
After excluding pathological slices with low image quality and those lacking mutation gene labels, the final dataset includes slides from Glioblastoma and Lower-Grade Glioma (GBMLGG; Num=1022), Breast Cancer (BRCA; Num=1001), Colorectal Cancer (CRC; Num=574), Uterine Corpus Endometrial Carcinoma (UCEC; Num=567), and Lung Adenocarcinoma (LUAD; Num=529). The detailed information of genetic mutation for all of the dataset shown in Table \ref{tab:1}.\par

\begin{table}[h]
    \centering
    \caption{The detailed information of genetic mutations in TCGA datasets}
    \label{tab:1}
    \renewcommand{\arraystretch}{1.2}
    \setlength{\tabcolsep}{3pt}
    \resizebox{\textwidth}{!}{
    \begin{tabular}{cccccccccccc}
        \toprule
        \multirow{3}{*}{\textbf{TCGA-GBMLGG}} & &\textbf{IDH1} & \textbf{TP53} & \textbf{ATRX} & \textbf{CIC} & \textbf{TTN} & \textbf{PIK3CA} & \textbf{EGFR} & \textbf{MUC16} & \textbf{FUBP1} & \textbf{PTEN} \\
        \cline{2-12}
        & Mutation & 681 & 420 & 310 & 210 & 127 & 123 & 110 & 99 & 91 & 80 \\
        & Wild Type & 341 & 602 & 712 & 812 & 895 & 899 & 912 & 923 & 931 & 942 \\
        \midrule
        \multirow{3}{*}{\textbf{TCGA-BRCA}} & &\textbf{PIK3CA} & \textbf{TP53} & \textbf{TTN} & \textbf{CDH1} & \textbf{GATA3} & \textbf{MUC16} & \textbf{KMT2C} & \textbf{MAP3K1} & \textbf{FLG} & \textbf{SYNE1} \\
        \cline{2-12}
        & Mutation & 355 & 347 & 184 & 126 & 122 & 111 & 94 & 91 & 67 & 66 \\
        & Wild Type & 646 & 654 & 817 & 875 & 879 & 890 & 907 & 910 & 934 & 935 \\
        \midrule
        \multirow{3}{*}{\textbf{TCGA-CRC}} & &\textbf{APC} & \textbf{TP53} & \textbf{KRAS} & \textbf{PIK3CA} & \textbf{FBXW7} & \textbf{SMAD4} & \textbf{AMER1} & \textbf{ARID1A} & \textbf{BRAF} & \textbf{ATM} \\
        \cline{2-12}
        & Mutation & 422 & 345 & 242 & 145 & 100 & 72 & 71 & 70 & 68 & 67 \\
        & Wild Type & 152 & 229 & 332 & 429 & 474 & 502 & 503 & 504 & 506 & 507 \\
        \midrule
        \multirow{3}{*}{\textbf{TCGA-UCEC}} & &\textbf{PTEN} & \textbf{PIK3CA} & \textbf{ARID1A} & \textbf{TP53} & \textbf{PIK3R1} & \textbf{KMT2D} & \textbf{CTNNB1} & \textbf{CTCF} & \textbf{ZFHX3} & \textbf{KMT2B} \\
        \cline{2-12}
        & Mutation & 344 & 264 & 259 & 191 & 175 & 157 & 144 & 138 & 125 & 122 \\
        & Wild Type & 185 & 265 & 270 & 338 & 354 & 372 & 385 & 391 & 404 & 407 \\
        \midrule
        \multirow{3}{*}{\textbf{TCGA-LUAD}} & &\textbf{BRAF} & \textbf{EGFR} & \textbf{MGA} & \textbf{KRAS} & \textbf{KEAP1} & \textbf{NF1} & \textbf{STK11} & \textbf{TP53} & \textbf{SMARCA4} & \textbf{RBM10} \\
        \cline{2-12}
        & Mutation & 256 & 134 & 103 & 82 & 67 & 65 & 55 & 41 & 40 & 39 \\
        & Wild Type & 273 & 395 & 426 & 447 & 462 & 464 & 474 & 488 & 489 & 490 \\ 
        \bottomrule
    \end{tabular}
    }
\end{table}

For each slide, we applied the preprocessing method from CLAM \cite{lu2021data}: removing the background at a $20\times$ magnification level and extracting non-overlapping patches of size $256\times256$ pixels.\par
At the feature extraction stage, for each non-overlapping patch of size $256\times256$ pixels obtained from WSIs, we extract 2048-dimensional features utilizing the ResNet-50 backbone network pre-trained by the self-supervised RetCCL \cite{wang2023retccl} model, and subsequently regularized the data within [-1,1] range. During the training phase of the M4 model, we employed the Adam \cite{kingma2014adam} optimizer with a learning rate of 0.0001 to train the model weights. Due to the generation of a large number of patches from WSIs, we used a batch size of 1. The dataset was divided into a training set and a test set in a 4:1 ratio. In the training set, we applied 5-fold cross-validation and reported the average results on the test set as the final outcomes. The evaluation metric for the M4 model in the gene mutation task was the area under the curve (AUC).\par



\subsection{Overall Experiments Performance}
We tested M4 performance on five TCGA datasets by classifying multiple genetic mutations simultaneously. And we select six single-task methods as comparison with our proposed M4 model, which are traditional instance-level MIL methods, including 1) Mean-Pooling; 2) Max-Pooling; 3) attention-based method AMIL \cite{chen2022pan}; 4) Transformer-based method TransMIL \cite{shao2021transmil}; 5) dual-stream architecture-based method DSMIL \cite{li2021dual}; 6) double-tier MIL framework DTFD-MIL \cite{zhang2022dtfd}.  For all methods, we employ the same feature extraction method to construct WSI bags,  and maintain consistency in hyperparameters and loss functions throughout the training process. And all models are trained and tested using the same training sets and test sets. The results of all other methods come from experiments conducted using their official codes under the same settings.\par
\begin{figure*}[!h]
  \centering
  \includegraphics[width=1\linewidth]{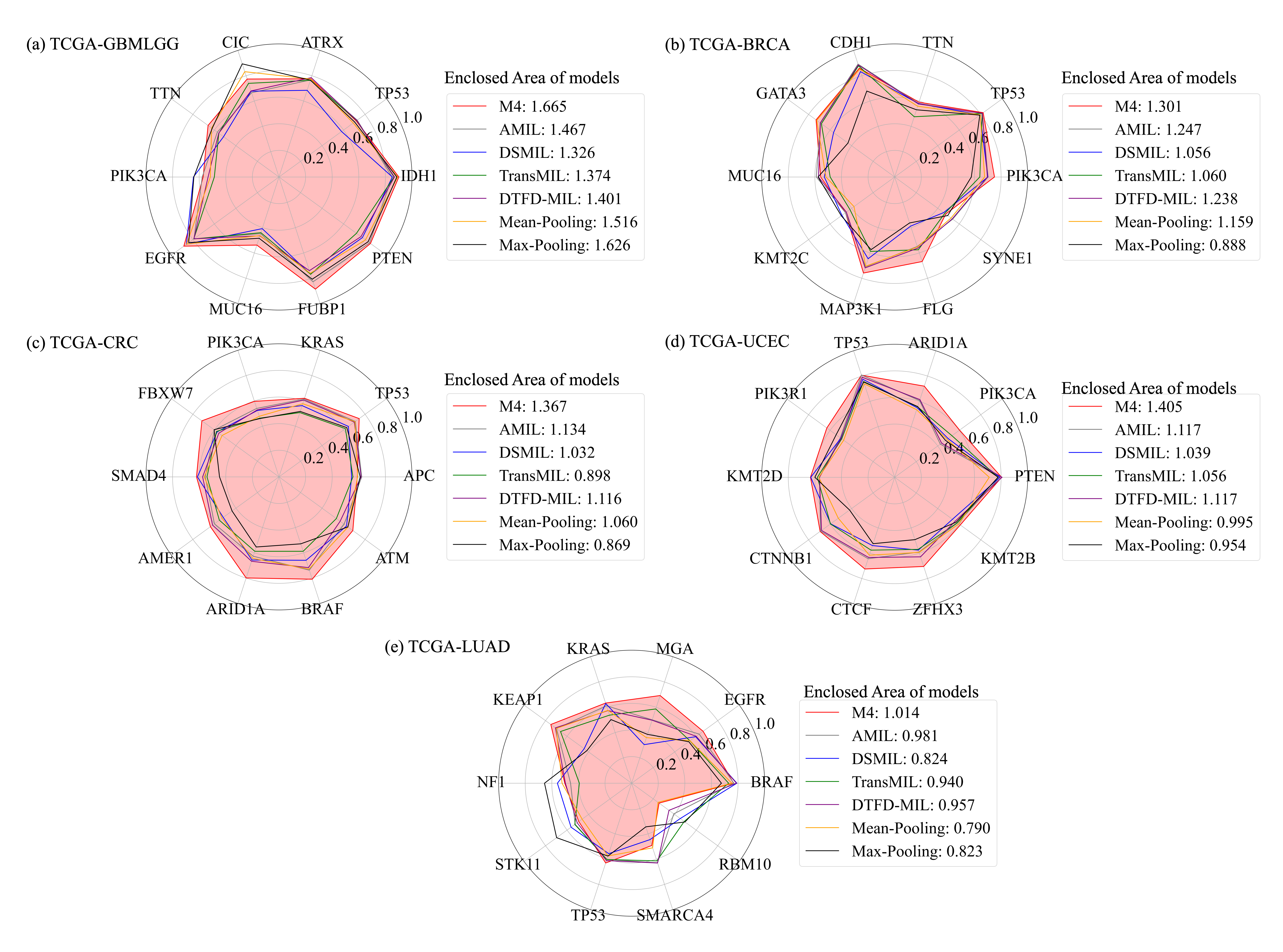}
  \caption{Radar charts of AUC performance of various models on five TCGA datasets. The red line represents the prediction results of the proposed model M4. The values in the legend represent the area enclosed by the radar charts.}
  \label{fig_3}
\end{figure*}

The average AUC for predicting 10 genetic mutations across five TCGA datasets is presented in Table \ref{tab:2}. Our M4 model achieved the best performance with an average prediction AUC of 0.677, which demonstrates consistent superiority over single-task models across five TCGA datasets. This is more evident in the TCGA-CRC and TCGA-UCEC  datasets, where M4 largely improved the average AUC by 6\% and  7\%, respectively. The detailed performance of each genetic mutation is illustrated in Figure \ref{fig_3} using radar charts, where each point represents the AUC prediction results for that gene, and the larger the area enclosed by the radar chart, the better the overall performance of the model across the 10 gene mutation tasks. The radar chart shows that our M4 method surpasses the single-task performance on all ten genetic mutation prediction tasks on the TCGA-CRC and TCGA-UCEC  datasets.  Specifically, M4 can improve the low-frequently mutated genes prediction, such as the AUC of MUC16 and FUBP1 were improved by 5.4\% and 5.8\% respectively in TCGA-GBMLGG (Figure 3(a)), and MAP3K1 and FLG were improved by 4\% and 9.3\% respectively in TCGA-BRCA dataset(Figure 3(b). This result supports our inference that  M4 can leverage the interrelationships between different genes to improve the prediction performance when predicting multiple genetic mutations simultaneously,.\par
\begin{table}[h]
    \centering
    \caption{\footnotesize Model AUC performance on 5 TCGA datasets. Best performance in \textbf{bold}, second best \underline{underlined}. For M4 model, the number of experts is five. The experimental results are the average AUC from five folds.}
    \resizebox{\linewidth}{!}{
    \label{tab:2}
    \renewcommand{\arraystretch}{1.2}
    \begin{tabular}{ccccccc}
        \toprule
        \textbf{Method} & \textbf{TCGA-GBMLGG} & \textbf{TCGA-BRCA} & \textbf{TCGA-CRC} & \textbf{TCGA-UCEC} & \textbf{TCGA-LUAD} & \textbf{Average}\\
        \midrule
        Mean-Pooling & 0.722 & 0.628 & 0.599 & 0.584 & 0.525 & 0.612\\
        Max-Pooling & \underline{0.747} & 0.551 & 0.543 & 0.572 & 0.528 & 0.588\\
        AMIL(\cite{chen2022pan}) & 0.707 & \underline{0.652} & \underline{0.621} & 0.621 & \underline{0.582} & \underline{0.636}\\
        DSMIL\cite{li2021dual} & 0.676 & 0.598 & 0.593 & 0.599 & 0.534 & 0.600\\
        TransMIL\cite{shao2021transmil} & 0.686 & 0.602 & 0.553 & 0.602 & 0.568 & 0.602\\
        DTFD-MIL\cite{zhang2022dtfd} & 0.691 & 0.650 & 0.616 & \underline{0.622} & 0.576 & 0.631\\
        M4(ours) & \textbf{0.756} & \textbf{0.667} & \textbf{0.682} & \textbf{0.693} & \textbf{0.588} & \textbf{0.677}\\
        \bottomrule
    \end{tabular}
    }
\end{table}
This result is consistent with our inference that when predicting multiple genetic mutations simultaneously, M4 can learn the interrelationships between different genes to improve the prediction performance.\par

\subsection{Ablation Study}
During our exploration of multi-task learning for WSI analysis, we designed and evaluated various strategies, such as utilizing AMIL to aggregate patch-level features and integrate them into MMoE. Though some of these models show superior performance over single-task models, they are limited by the learning capacity of the experts. Therefore, we designed MP-AMIL to enhance the spatial and channel multi-proxy learning capabilities of the experts network. To adapt to MP-AMIL, we further proposed a multi-proxy gate network MP-Gate, which ultimately led to the development of our M4 model. The ablation study performance is shown in Table \ref{tab:3}. The results demonstrate that MP-AMIL and corresponding MP-Gate can enhance the task performance of MIL in the MMoE framework. \par
\begin{table}[t]
    \centering
    \caption{\footnotesize Ablation model AUC performance on five datasets. Best performance in \textbf{bold}, second best \underline{underlined}. MMoE+AMIL(expert) denotes using AMIL as the expert network, while MMoE+MP-AMIL(expert) denotes using our proposed MP-AMIL as the expert network. M4 is our proposed model. The number of experts for all models in this experiment is set to 5.}
    \resizebox{\linewidth}{!}{
    \label{tab:3}
    \renewcommand{\arraystretch}{1.2}
    \begin{tabular}{ccccccc}
        \toprule
        \textbf{Method} & \textbf{TCGA-GBMLGG} & \textbf{TCGA-BRCA} & \textbf{TCGA-CRC} & \textbf{TCGA-UCEC} & \textbf{TCGA-LUAD} & \textbf{Average}\\
        \midrule
        AMIL       & 0.707 & 0.652 & 0.621 & 0.621 & \underline{0.582} & 0.636\\
        MMoE+AMIL(expert) & \underline{0.759} & 0.601 & 0.662 & 0.682 & 0.566 & 0.654\\
        MMoE+MP-AMIL(expert)& \textbf{0.763} & \underline{0.665} & \underline{0.675} & \underline{0.687} & 0.570 & \underline{0.672}\\
        \textbf{MMoE+MP-AMIL+MP-Gate(ours-M4)} & 0.756 & \textbf{0.667} & \textbf{0.682} & \textbf{0.693} & \textbf{0.588} & \textbf{0.677}\\
        \bottomrule
    \end{tabular}
    }
\end{table}

\begin{figure*}[h]
  \centering
  \includegraphics[width=1\linewidth]{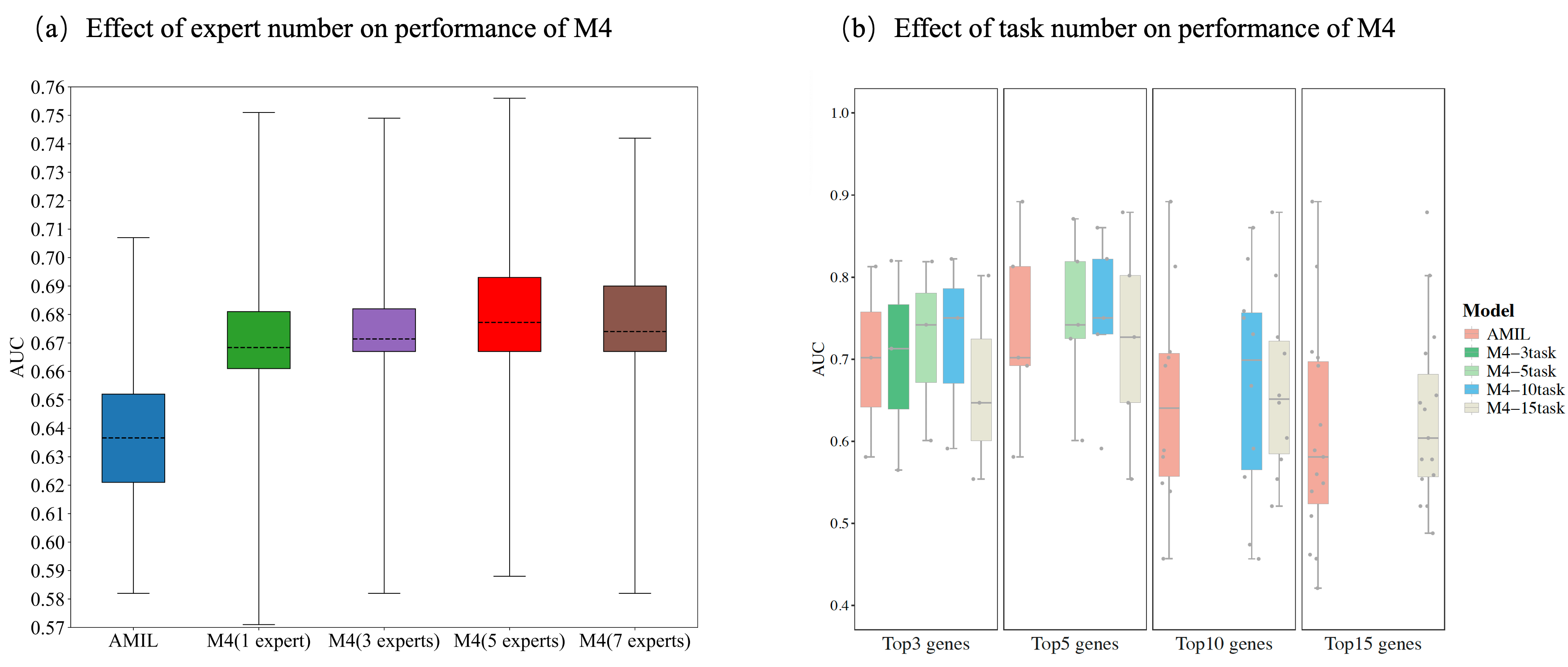}
  \caption{Box plots highlighting the AUC performance of our proposed M4 model under different numbers of experts and different numbers of tasks.}
  \label{fig_4}
\end{figure*}

We also examine the influence of the number of experts on the performance of the M4 framework. Figure\ref{fig_4}(a) illustrates the AUC scores for different numbers of experts in M4 respectively. The number of tasks for all M4 models set to 10. All experiments were conducted under the same conditions as in the previous experiments, except for the number of experts for comparisons. The results show that 1) Generally the M4 framework achieves a significant increase compared to the single-task model AMIL, irrespective of the number of experts. 2) The overall performance difference across various numbers of experts of the M4 framework is not obvious, with the 5-expert model showing a slightly higher performance.\par

To validate the performance of our proposed M4 model under different task numbers. Taking BRCA as an example, we select the top-$k$ genes ($k$=3, 5, 10, 15) based on mutation frequency for model training, validation and testing. The number of experts for all M4 models set to 5. To ensure the comparability of model results, we selected the same genes under different numbers of tasks for comparison. The test results are shown in Figure\ref{fig_4}(b). It can be observed that, among the top 3 genes, all M4 models surpass the single-task model AMIL except for the M4 model with 15 tasks.  The M4 model with 10 tasks achieves the best performance. Additionally, for the top5, top10, and top15 genes, the AUC performance of the M4 models consistently exceeds that of AMIL. Moreover, it is evident that for the M4 model, as the number of tasks increases, the AUC performance for highly mutated genes initially improves. However, when the number of tasks becomes larger, the AUC performance for highly mutated genes compromises towards that of low-mutated genes. Consequently, at this stage, the M4 model demonstrates a more pronounced improvement in the prediction performance for genes with low mutation frequencies.


\subsection{Heatmap Visualization}
\begin{figure*}[t]
  \centering
  \includegraphics[width=1\linewidth]{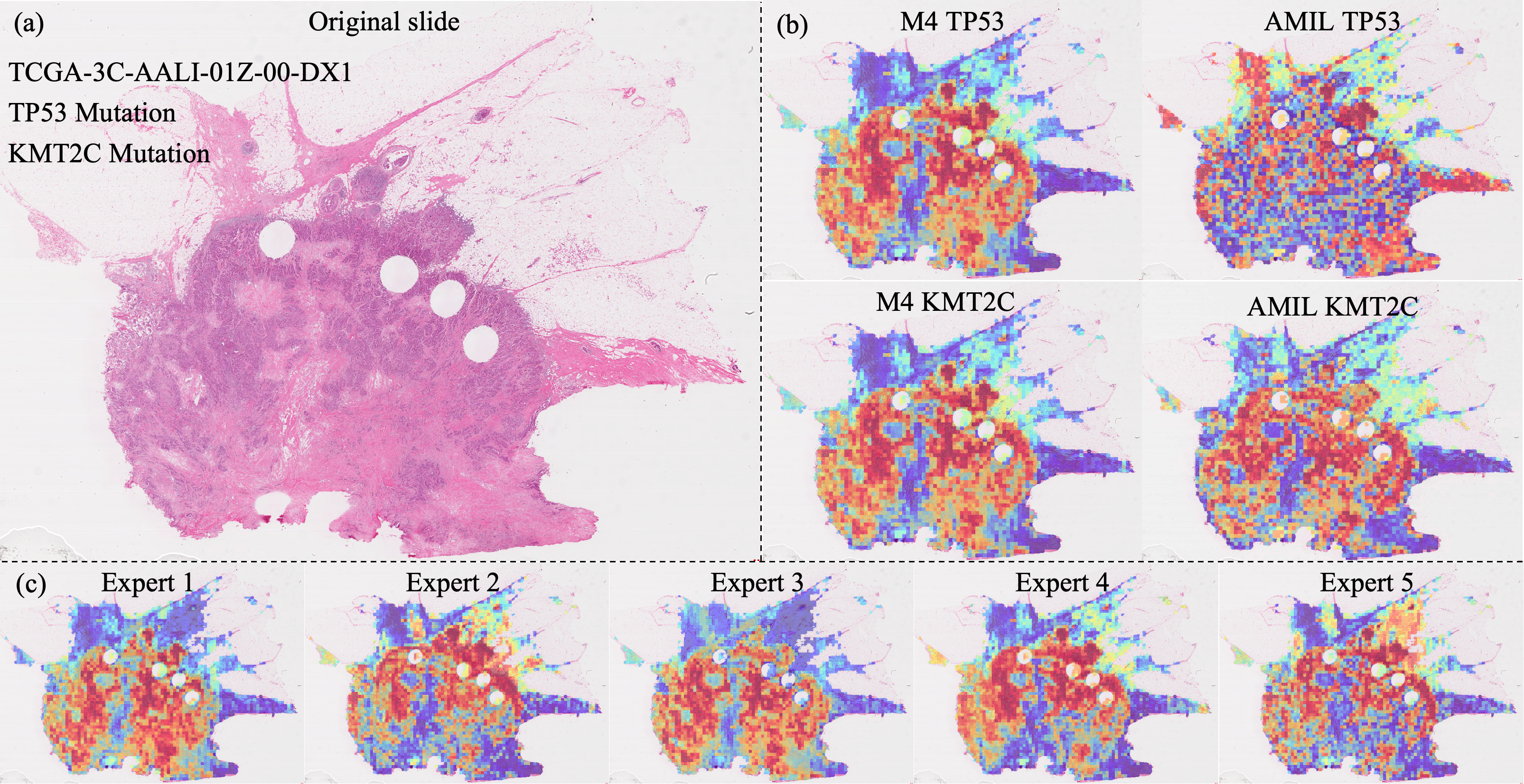}
  \caption{Heatmap Visualization for M4. (a) Original slide. In original slide, the areas with purple are more likely to be the tumor regions. In the other column. (b) Heatmaps of the slide by single-task AMIL attention scores and by M4 model, respectively. Warm colors indicate higher probabilities to be the region of interest for the corresponding locations. (c) Heatmap of different experts by M4 model.}
  \label{fig_5}
\end{figure*}

Based on the heatmap, we can analyze whether the model is focusing on the tumor region and whether each expert in the model is playing an important role. We investigate the ability of our M4 model in handling the intricate phenotypes present in WSIs by visualizing the attention heatmaps generated by different experts and different task. The comparison with the AMIL single-task  method is illustrated in Figure \ref{fig_5}.  Figure \ref{fig_5}(a) is the original slide. Figure \ref{fig_5}(b) is task related heatmaps for this slide. Compared to AMIL, M4 model consistently identifies potential tumor regions. Figure \ref{fig_5}(c) shows each expert attention heatmap. As expected, the regions of focus for each expert, indicating that different experts can learn different tumor information.\par

\section{Discussion}

To effectively predict multiple genetic mutations simultaneously in pathology images, we propose the M4 model. Our model combines Multi-gate Mixture-of-Experts (MMoE) with Multiple Instance Learning (MIL) and achieves improved genetic mutation prediction performance across multiple cancer datasets. As illustrated in Figure \ref{fig_3}, we compare the prediction performance of M4 with other MIL methods on different genetic mutations across five datasets, demonstrating the superior performance of M4.\par

To further investigate the effectiveness of each module in M4, we conducted ablation experiments. The results indicate that merely integrating AMIL as the expert network into MMoE does not achieve satisfactory prediction performance. Building on AMIL, the introduction of MP-AMIL, which incorporates multi-proxy information, enhances the performance of MIL when combined with MMoE. Ultimately, by adapting the gate network with MP-AMIL, we proposed the M4 model, which further improves prediction performance.

We also compared the interpretability heatmaps generated by M4 and AMIL, as shown in Figure \ref{fig_4}. Overall, M4 focuses more on suspected tumor regions. Additionally, we visualized the weight heatmaps of each expert, further confirming that in M4, each expert learns different content, indicating that M4, compared to AMIL, can capture more tumor features from WSIs.\par

However, our study has certain limitations and areas for improvement. The current M4 model focuses more on common features among genes and does not explicitly learn the unique characteristics of individual genes, which may lead to sub-optimal performance when handling conflicting gene features. Therefore, incorporating the unique features of individual genes is a promising direction for future research.

\section{Conclusion}

In this work, we introduce the M4 framework designed for multi-task learning in WSI analysis, aiming to meet the growing demand in the CPath field. Specifically, we draw inspiration from the extensively used Multi-gate Mixture-of-experts architecture and adapt it to the Multiple instance learning framework. Furthermore, we construct multi-proxy expert networks and gating networks for sufficiently and comprehensively modeling correlations between patches of different scopes within WSIs. The M4 model surpasses current state-of-the-art single-task MIL models when evaluated on five TCGA datasets for genetic mutation prediction tasks. The M4 model also demonstrates performance improvement with a limited number of positive samples. The visualization process of the heatmaps illustrates that M4 is capable of more broadly focusing on the tumor regions within the slide tissue. We believe our work is an important advancement in WSI analysis, and expect the M4 to serve as a valuable tool for more MIL use cases.
\section{Acknowledge}
This research was funded by the Zhejiang Leading Innovation and Entrepreneurship Team (No. 2022R01006).
\bibliographystyle{unsrt}
\bibliography{M4-main}

\begin{thebibliography}{10}

\bibitem{shmatko2022artificial}
Artem Shmatko, Narmin Ghaffari~Laleh, Moritz Gerstung, and Jakob~Nikolas Kather.
\newblock Artificial intelligence in histopathology: enhancing cancer research and clinical oncology.
\newblock {\em Nature cancer}, 3(9):1026--1038, 2022.

\bibitem{song2023artificial}
Andrew~H Song, Guillaume Jaume, Drew~FK Williamson, Ming~Y Lu, Anurag Vaidya, Tiffany~R Miller, and Faisal Mahmood.
\newblock Artificial intelligence for digital and computational pathology.
\newblock {\em Nature Reviews Bioengineering}, 1(12):930--949, 2023.

\bibitem{li2022comprehensive}
Xintong Li, Chen Li, Md~Mamunur Rahaman, Hongzan Sun, Xiaoqi Li, Jian Wu, Yudong Yao, and Marcin Grzegorzek.
\newblock A comprehensive review of computer-aided whole-slide image analysis: from datasets to feature extraction, segmentation, classification and detection approaches.
\newblock {\em Artificial Intelligence Review}, 55(6):4809--4878, 2022.

\bibitem{zhang2019pathologist}
Zizhao Zhang, Pingjun Chen, Mason McGough, Fuyong Xing, Chunbao Wang, Marilyn Bui, Yuanpu Xie, Manish Sapkota, Lei Cui, Jasreman Dhillon, et~al.
\newblock Pathologist-level interpretable whole-slide cancer diagnosis with deep learning.
\newblock {\em Nature Machine Intelligence}, 1(5):236--245, 2019.

\bibitem{skrede2020deep}
Ole-Johan Skrede, Sepp De~Raedt, Andreas Kleppe, Tarjei~S Hveem, Knut Liest{\o}l, John Maddison, Hanne~A Askautrud, Manohar Pradhan, John~Arne Nesheim, Fritz Albregtsen, et~al.
\newblock Deep learning for prediction of colorectal cancer outcome: a discovery and validation study.
\newblock {\em The Lancet}, 395(10221):350--360, 2020.

\bibitem{li2021deep}
Fengling Li, Yongquan Yang, Yani Wei, Ping He, Jie Chen, Zhongxi Zheng, and Hong Bu.
\newblock Deep learning-based predictive biomarker of pathological complete response to neoadjuvant chemotherapy from histological images in breast cancer.
\newblock {\em Journal of translational medicine}, 19:1--13, 2021.

\bibitem{kather2019deep}
Jakob~Nikolas Kather, Alexander~T Pearson, Niels Halama, Dirk J{\"a}ger, Jeremias Krause, Sven~H Loosen, Alexander Marx, Peter Boor, Frank Tacke, Ulf~Peter Neumann, et~al.
\newblock Deep learning can predict microsatellite instability directly from histology in gastrointestinal cancer.
\newblock {\em Nature medicine}, 25(7):1054--1056, 2019.

\bibitem{fu2020pan}
Yu~Fu, Alexander~W Jung, Ramon~Vi{\~n}as Torne, Santiago Gonzalez, Harald V{\"o}hringer, Artem Shmatko, Lucy~R Yates, Mercedes Jimenez-Linan, Luiza Moore, and Moritz Gerstung.
\newblock Pan-cancer computational histopathology reveals mutations, tumor composition and prognosis.
\newblock {\em Nature cancer}, 1(8):800--810, 2020.

\bibitem{qu2021genetic}
Hui Qu, Mu~Zhou, Zhennan Yan, He~Wang, Vinod~K Rustgi, Shaoting Zhang, Olivier Gevaert, and Dimitris~N Metaxas.
\newblock Genetic mutation and biological pathway prediction based on whole slide images in breast carcinoma using deep learning.
\newblock {\em NPJ precision oncology}, 5(1):87, 2021.

\bibitem{tsai2023histopathology}
Pei-Chen Tsai, Tsung-Hua Lee, Kun-Chi Kuo, Fang-Yi Su, Tsung-Lu~Michael Lee, Eliana Marostica, Tomotaka Ugai, Melissa Zhao, Mai~Chan Lau, and Juha~P V{\"a}yrynen.
\newblock Histopathology images predict multi-omics aberrations and prognoses in colorectal cancer patients.
\newblock {\em Nature communications}, 14(1):2102, 2023.

\bibitem{abdel2023comprehensive}
Heba Abdel-Nabi, Mostafa Ali, Arafat Awajan, Mohammad Daoud, Rami Alazrai, Ponnuthurai~N Suganthan, and Talal Ali.
\newblock A comprehensive review of the deep learning-based tumor analysis approaches in histopathological images: segmentation, classification and multi-learning tasks.
\newblock {\em Cluster Computing}, 26(5):3145--3185, 2023.

\bibitem{kumar2020whole}
Neeta Kumar, Ruchika Gupta, and Sanjay Gupta.
\newblock Whole slide imaging (wsi) in pathology: current perspectives and future directions.
\newblock {\em Journal of digital imaging}, 33(4):1034--1040, 2020.

\bibitem{hu2023state}
Weiming Hu, Xintong Li, Chen Li, Rui Li, Tao Jiang, Hongzan Sun, Xinyu Huang, Marcin Grzegorzek, and Xiaoyan Li.
\newblock A state-of-the-art survey of artificial neural networks for whole-slide image analysis: from popular convolutional neural networks to potential visual transformers.
\newblock {\em Computers in Biology and Medicine}, 161:107034, 2023.

\bibitem{hou2022h}
Wentai Hou, Lequan Yu, Chengxuan Lin, Helong Huang, Rongshan Yu, Jing Qin, and Liansheng Wang.
\newblock H\^{} 2-mil: exploring hierarchical representation with heterogeneous multiple instance learning for whole slide image analysis.
\newblock In {\em Proceedings of the AAAI conference on artificial intelligence}, volume~36, pages 933--941, 2022.

\bibitem{kanavati2020weakly}
Fahdi Kanavati, Gouji Toyokawa, Seiya Momosaki, Michael Rambeau, Yuka Kozuma, Fumihiro Shoji, Koji Yamazaki, Sadanori Takeo, Osamu Iizuka, and Masayuki Tsuneki.
\newblock Weakly-supervised learning for lung carcinoma classification using deep learning.
\newblock {\em Scientific reports}, 10(1):9297, 2020.

\bibitem{wang2022scl}
Xiyue Wang, Jinxi Xiang, Jun Zhang, Sen Yang, Zhongyi Yang, Ming-Hui Wang, Jing Zhang, Wei Yang, Junzhou Huang, and Xiao Han.
\newblock Scl-wc: Cross-slide contrastive learning for weakly-supervised whole-slide image classification.
\newblock {\em Advances in neural information processing systems}, 35:18009--18021, 2022.

\bibitem{wang2021transpath}
Xiyue Wang, Sen Yang, Jun Zhang, Minghui Wang, Jing Zhang, Junzhou Huang, Wei Yang, and Xiao Han.
\newblock Transpath: Transformer-based self-supervised learning for histopathological image classification.
\newblock In {\em Medical Image Computing and Computer Assisted Intervention--MICCAI 2021: 24th International Conference, Strasbourg, France, September 27--October 1, 2021, Proceedings, Part VIII 24}, pages 186--195. Springer, 2021.

\bibitem{lazard2023giga}
Tristan Lazard, Marvin Lerousseau, Etienne Decenci{\`e}re, and Thomas Walter.
\newblock Giga-ssl: Self-supervised learning for gigapixel images.
\newblock In {\em Proceedings of the IEEE/CVF Conference on Computer Vision and Pattern Recognition}, pages 4305--4314, 2023.

\bibitem{schmidt2022efficient}
Arne Schmidt, Julio Silva-Rodr{\'\i}guez, Rafael Molina, and Valery Naranjo.
\newblock Efficient cancer classification by coupling semi supervised and multiple instance learning.
\newblock {\em IEEE Access}, 10:9763--9773, 2022.

\bibitem{xie2020beyond}
Chensu Xie, Hassan Muhammad, Chad~M Vanderbilt, Raul Caso, Dig Vijay~Kumar Yarlagadda, Gabriele Campanella, and Thomas~J Fuchs.
\newblock Beyond classification: Whole slide tissue histopathology analysis by end-to-end part learning.
\newblock In {\em MIDL}, pages 843--856, 2020.

\bibitem{ilse2018attention}
Maximilian Ilse, Jakub Tomczak, and Max Welling.
\newblock Attention-based deep multiple instance learning.
\newblock In {\em International conference on machine learning}, pages 2127--2136. PMLR, 2018.

\bibitem{chen2022pan}
Richard~J Chen, Ming~Y Lu, Drew~FK Williamson, Tiffany~Y Chen, Jana Lipkova, Zahra Noor, Muhammad Shaban, Maha Shady, Mane Williams, and Bumjin Joo.
\newblock Pan-cancer integrative histology-genomic analysis via multimodal deep learning.
\newblock {\em Cancer Cell}, 40(8):865--878, 2022.

\bibitem{shao2021transmil}
Zhuchen Shao, Hao Bian, Yang Chen, Yifeng Wang, Jian Zhang, and Xiangyang Ji.
\newblock Transmil: Transformer based correlated multiple instance learning for whole slide image classification.
\newblock {\em Advances in neural information processing systems}, 34:2136--2147, 2021.

\bibitem{li2021dt}
Hang Li, Fan Yang, Yu~Zhao, Xiaohan Xing, Jun Zhang, Mingxuan Gao, Junzhou Huang, Liansheng Wang, and Jianhua Yao.
\newblock Dt-mil: deformable transformer for multi-instance learning on histopathological image.
\newblock In {\em Medical Image Computing and Computer Assisted Intervention--MICCAI 2021: 24th International Conference, Strasbourg, France, September 27--October 1, 2021, Proceedings, Part VIII 24}, pages 206--216. Springer, 2021.

\bibitem{gao2023semi}
Zeyu Gao, Bangyang Hong, Yang Li, Xianli Zhang, Jialun Wu, Chunbao Wang, Xiangrong Zhang, Tieliang Gong, Yefeng Zheng, and Deyu Meng.
\newblock A semi-supervised multi-task learning framework for cancer classification with weak annotation in whole-slide images.
\newblock {\em Medical Image Analysis}, 83:102652, 2023.

\bibitem{liu2023multi}
Jianxin Liu, Rongjun Ge, Peng Wan, Qi~Zhu, Daoqiang Zhang, and Wei Shao.
\newblock Multi-task multi-instance learning for jointly diagnosis and prognosis of early-stage breast invasive carcinoma from whole-slide pathological images.
\newblock In {\em International Conference on Information Processing in Medical Imaging}, pages 145--157. Springer, 2023.

\bibitem{ma2018modeling}
Jiaqi Ma, Zhe Zhao, Xinyang Yi, Jilin Chen, Lichan Hong, and Ed~H Chi.
\newblock Modeling task relationships in multi-task learning with multi-gate mixture-of-experts.
\newblock In {\em Proceedings of the 24th ACM SIGKDD international conference on knowledge discovery \& data mining}, pages 1930--1939, 2018.

\bibitem{zheng2022survey}
Yong Zheng and David~Xuejun Wang.
\newblock A survey of recommender systems with multi-objective optimization.
\newblock {\em Neurocomputing}, 474:141--153, 2022.

\bibitem{campanella2019clinical}
Gabriele Campanella, Matthew~G Hanna, Luke Geneslaw, Allen Miraflor, Vitor Werneck Krauss~Silva, Klaus~J Busam, Edi Brogi, Victor~E Reuter, David~S Klimstra, and Thomas~J Fuchs.
\newblock Clinical-grade computational pathology using weakly supervised deep learning on whole slide images.
\newblock {\em Nature medicine}, 25(8):1301--1309, 2019.

\bibitem{lerousseau2020weakly}
Marvin Lerousseau, Maria Vakalopoulou, Marion Classe, Julien Adam, Enzo Battistella, Alexandre Carr{\'e}, Th{\'e}o Estienne, Th{\'e}ophraste Henry, Eric Deutsch, and Nikos Paragios.
\newblock Weakly supervised multiple instance learning histopathological tumor segmentation.
\newblock In {\em Medical Image Computing and Computer Assisted Intervention--MICCAI 2020: 23rd International Conference, Lima, Peru, October 4--8, 2020, Proceedings, Part V 23}, pages 470--479. Springer, 2020.

\bibitem{xu2019camel}
Gang Xu, Zhigang Song, Zhuo Sun, Calvin Ku, Zhe Yang, Cancheng Liu, Shuhao Wang, Jianpeng Ma, and Wei Xu.
\newblock Camel: A weakly supervised learning framework for histopathology image segmentation.
\newblock In {\em Proceedings of the IEEE/CVF International Conference on computer vision}, pages 10682--10691, 2019.

\bibitem{li2021dual}
Bin Li, Yin Li, and Kevin~W Eliceiri.
\newblock Dual-stream multiple instance learning network for whole slide image classification with self-supervised contrastive learning.
\newblock In {\em Proceedings of the IEEE/CVF conference on computer vision and pattern recognition}, pages 14318--14328, 2021.

\bibitem{zhang2022dtfd}
Hongrun Zhang, Yanda Meng, Yitian Zhao, Yihong Qiao, Xiaoyun Yang, Sarah~E Coupland, and Yalin Zheng.
\newblock Dtfd-mil: Double-tier feature distillation multiple instance learning for histopathology whole slide image classification.
\newblock In {\em Proceedings of the IEEE/CVF conference on computer vision and pattern recognition}, pages 18802--18812, 2022.

\bibitem{tang2020progressive}
Hongyan Tang, Junning Liu, Ming Zhao, and Xudong Gong.
\newblock Progressive layered extraction (ple): A novel multi-task learning (mtl) model for personalized recommendations.
\newblock In {\em Proceedings of the 14th ACM Conference on Recommender Systems}, pages 269--278, 2020.

\bibitem{zhang2022multi}
Qihua Zhang, Junning Liu, Yuzhuo Dai, Yiyan Qi, Yifan Yuan, Kunlun Zheng, Fan Huang, and Xianfeng Tan.
\newblock Multi-task fusion via reinforcement learning for long-term user satisfaction in recommender systems.
\newblock In {\em Proceedings of the 28th ACM SIGKDD conference on knowledge discovery and data mining}, pages 4510--4520, 2022.

\bibitem{caruana1997multitask}
Rich Caruana.
\newblock Multitask learning.
\newblock {\em Machine learning}, 28:41--75, 1997.

\bibitem{reisenbuchler2022local}
Daniel Reisenb{\"u}chler, Sophia~J Wagner, Melanie Boxberg, and Tingying Peng.
\newblock Local attention graph-based transformer for multi-target genetic alteration prediction.
\newblock In {\em International Conference on Medical Image Computing and Computer-Assisted Intervention}, pages 377--386. Springer, 2022.

\bibitem{jacobs1991adaptive}
Robert~A Jacobs, Michael~I Jordan, Steven~J Nowlan, and Geoffrey~E Hinton.
\newblock Adaptive mixtures of local experts.
\newblock {\em Neural computation}, 3(1):79--87, 1991.

\bibitem{zhao2019multiple}
Jiejie Zhao, Bowen Du, Leilei Sun, Fuzhen Zhuang, Weifeng Lv, and Hui Xiong.
\newblock Multiple relational attention network for multi-task learning.
\newblock In {\em Proceedings of the 25th ACM SIGKDD international conference on knowledge discovery \& Data Mining}, pages 1123--1131, 2019.

\bibitem{hou2016patch}
Le~Hou, Dimitris Samaras, Tahsin~M Kurc, Yi~Gao, James~E Davis, and Joel~H Saltz.
\newblock Patch-based convolutional neural network for whole slide tissue image classification.
\newblock In {\em Proceedings of the IEEE conference on computer vision and pattern recognition}, pages 2424--2433, 2016.

\bibitem{wang2023retccl}
Xiyue Wang, Yuexi Du, Sen Yang, Jun Zhang, Minghui Wang, Jing Zhang, Wei Yang, Junzhou Huang, and Xiao Han.
\newblock Retccl: Clustering-guided contrastive learning for whole-slide image retrieval.
\newblock {\em Medical image analysis}, 83:102645, 2023.

\bibitem{Chollet2017xception}
Fran{\c{c}}ois Chollet.
\newblock Xception: Deep learning with depthwise separable convolutions.
\newblock In {\em Proceedings of the IEEE conference on computer vision and pattern recognition}, pages 1251--1258, 2017.

\bibitem{lu2021data}
Ming~Y Lu, Drew~FK Williamson, Tiffany~Y Chen, Richard~J Chen, Matteo Barbieri, and Faisal Mahmood.
\newblock Data-efficient and weakly supervised computational pathology on whole-slide images.
\newblock {\em Nature biomedical engineering}, 5(6):555--570, 2021.

\bibitem{kingma2014adam}
Diederik~P Kingma and Jimmy Ba.
\newblock Adam: A method for stochastic optimization.
\newblock {\em arXiv preprint arXiv:1412.6980}, 2014.

\end{thebibliography}

\end{document}